# Mimicking Ensemble Learning with Deep Branched Networks

Byungju Kim, Youngsoo Kim, Yeakang Lee, Junmo Kim

Department of Electric Engineering, KAIST

{feidfoe, zamgyul, askhow, junmo.kim}@kaist.ac.kr,

## Abstract

This paper proposes a branched residual network for image classification. It is known that high-level features of deep neural network are more representative than lower-level features. By sharing the low-level features, the network can allocate more memory to high-level features. The upper layers of our proposed network are branched, so that it mimics the ensemble learning. By mimicking ensemble learning with single network, we have achieved better performance on ImageNet classification task.

## 1. Introduction

Deep networks, which are used in various fields in recent years, include many layers. In 2012, the AlexNet [1] structure, which uses only 8 layers, has greatly improved the existing image recognition rate, and after the rapid technological development and deepening of the network structure, more than 1000 layers of networks are announced [4].

This very deep network has the best performance in a variety of areas, but it requires a lot of memory and computation in learning and verification because of its architectural features with more than 1000 layers. In recent years, as the tendency to use deeper and deeper networks has become stronger, high-performance image recognizers are learned by using expensive equipment, and a recognizer that transcends human ability has appeared.

Accordingly, in this task, an image recognizer with high performance is developed. However, deeper network directly mean that the network contains more learnable parameters, so that the over-fitting problem could be severe.

To improve the performance of our proposed network, we have applied various data augmentation and label smoothing technique to train the branched residual network. We have used ImageNet [2] classification task to verify our method.

Rest of this paper is composed of four chapters. In the following chapter, we present the branched residual network for ImageNet classification task. In chapter 3, we propose label smoothing method. Then, we show the experimental results in chapter 4 and conclude this paper in chapter 5. In chapter 5, we have mentioned the limit and future work of our work. We have used validation images of ImageNet Large Scale Visual Recognition Challenge 2016 dataset for our experiments.

## 2. Branched Residual Network

In this paper, we have proposed branched residual network for efficient memory usage. Residual network [4] is deep neural network for image classification task proposed in [4]. Existing networks tend to show better performance as they have a deeper structure, but their depth is limited. In addition, when the network structure becomes deeper and deeper (> 20), the performance is deteriorated. It is because the gradient signal generated from the top layer vanishes as it passes through many layers. The gradient vanishing problem is caused by multiple reasons. Intuitive reason is that the synaptic weight of a deep network actually has very small value, and there are signals that disappears as it passes through the ReLU function and the pooling layer. By using shortcut connection, the residual network successfully conveys the error signal from the top to the lowest layer. This make the deeper network can perform better.

Deep networks extract new features by combining the features of the lower layers. Hence, the features extracted by upper layers are more complex and expressive shapes. Therefore, deep networks extract features close to the complete objects as they go to the upper layer. In the lower part of the network, extracted feature maps are mainly focused on edge and color information.

For example, in the case of car classification, color and edge information is extracted from the lower layer, while higher-level features such as wheel and window are extracted in the upper layer. The extracted information is combined to finally classify the image. If the network is learned to recognize a person's face, the features of eyes, nose, mouth, and the entire face are extracted from the upper layer. That is, the characteristics of the upper layer reflect the characteristics of the training data and the domain. On the other hand, low-level features such as edges are extracted at the lower layer without being greatly affected by learning data and its domain.

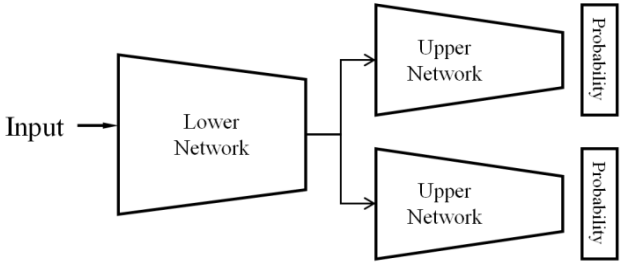

**Figure 1** Overall architecture of branched residual network

In this paper, we propose a network that shares the feature maps of the lower layers and learns the upper layers independently to utilize the memory efficiently. Figure 1 present the overall architecture of proposed branched residual network. It outputs two final probabilities that are generated independently. In real implementation, we have used residual network with 200 convolutional layers. It is composed of 66 residual blocks. We have branched the network after 39th residual block, so that the whole network is composed of 93 residual blocks.

## 3. Label Smoothing

Label smoothing technique, proposed in [3], is also referred as label-dropout. In conventional supervised classification task, given ground-truth label for each image is one-hot encoded vector. Deep neural network is trained to infer outputs similar to the ground-truth label. However, since the output of the network is calculated by softmax function, one-hot encoded label is unreachable goal. Instead of one-hot encoded ground-truth, the label smoothing technique suggests the ground-truth label as follow:

$$p_i = \begin{cases} 1 - \varepsilon + \frac{\varepsilon}{K} & if\ i\ is\ ground\ truth\ class \\ \frac{\varepsilon}{K} & otherwise \end{cases},$$

where K is the number of classes. This method can be interpreted as a method of regularizing the distribution of the predicted value from the model as a prior with a uniform distribution.

We can analyze the label smoothing method from a different viewpoint. Smoothed label can play a role as distilled knowledge [6]. In case of two classes are very similar, one-hot encoded vector always indicates the two classes are strictly different. However, with smoothed label, indicated difference between the two classes is smaller, so that the network has the potential to learn the distilled knowledge from the training data. By applying label smoothing technique, each branch of our proposed network performs better, so that the final performance has been improved significantly.

In the following chapter, we present our experimental results using ImageNet object classification task. Among the dataset, we have used validation set to verify our method.

## 4. Experimental Results

The performance of the trained network can vary not only by the structure of the network but also by the data used in training and by the method. Since the same images have used during the training repeatedly for several epochs, over-fitting problem may occur. Data augmentation is one of the methods for generalizing the network. Data augmentation, which is implemented in Caffe [5], includes random crop and horizontal flip. Random crop is a method that crops out some areas in a large image and uses it for training instead of whole image. It is a generalization technique to maintain the object ratio from being damaged when the image size is arbitrarily resized. Horizontal flip is a method of learning by reversing the image to the left and right with a probability of 0.5 during the training.

| Original image | Brightness noise |
|---|---|
| Saturation noise | Contrast noise |

**Table 1** The effect of color augmentation.

In addition to the existing generalization techniques, we also implemented PCA noise, color augmentation, batch shuffle, and color normalization. PCA noise is a method of calculating the first principal component, the second principal component, and the third principal component for each pixel of the entire training image and assigning noise to each vector. It is more effective to add noise similar to the distribution of real image than to add noise directly to RGB values. Color augmentation is a method of transforming an image by adding noise to the brightness, saturation, and contrast values of the image. Table 1 shows the effect of color augmentation. It maintains the semantic information while randomly transforming the image.

Batch shuffle randomly permutes the training image every epoch. This prevents the over-fitting problem by changing the configuration of the images included in single mini-batch every time. It is reported that batch shuffling significantly contributes the performance improvement. Finally, color normalization is a way of adjusting the pixel mean of the image to zero, as in batch normalization layer. The implementation of color normalization is straight forward since the operation is identical to the batch normalization.

Data augmentation does not deform the image with a predetermined value for each image. Randomness is the most important factor of data augmentation. It arbitrarily transforms the image during the training, so that it is impossible to perform this operation on the image in advance. Hence, though we have improved the performance through data augmentation, it requires more training time.

Our code is implemented using Torch7. We used four TitanX gpus to train each model.

|  | Branched ResNet | Branched ResNet-LS |
|---|---|---|
| Batch size | 128 | 128 |
| Total epoch | 95 | 95 |
| Base learning rate | 0.05 | 0.05 |
| Learning rate decay | ×0.1 every 30epoch | ×0.1 every 30epoch |
| Weight decay | 0.0001 | 0.0001 |
| Label smoothing | X | O |
| Classification error | 3.86 | 3.72 |

**Table 2** Our experimental results and hyper parameters to train the network

Table 2 shows the experimental result of our method. Branched ResNet represent the architecture depicted in figure 1, trained with one-hot encoded ground-truth vector. Branched ResNet-LS represents the model trained with label smoothing technique with same architecture.

The error rates in table 2 indicate the result of ensemble of two branches for the validation set of ImageNet classification data. The classification error is calculated based on top-5 error rate. As it is shown in table 2, Branched ResNet-LS outperforms the Branched ResNet. We have invested the performance of each branch, so that we could analyze the performance improvement.

|  | Branched ResNet | Branched ResNet-LS |
|---|---|---|
| Branch1 | 22.02 | 21.24 |
| Branch2 | 22.09 | 21.32 |
| Ensemble | 20.81 | 20.31 |
| Relative Improvement | 5.65 | 4.56 |

**Table 3** Experimental results on each branch and ensemble of branches in top-1 error rate without test augmentation.

Table 3 indicates the top-1 error rate of each branch and final result of the whole network without test augmentation. It shows that, the performance improvement of Branched ResNet-LS results from the performance gain on individual branches. On the other hand, the effect of ensemble is more significant in Branched ResNet.

## 5. Conclusion

The proposed branched residual network shares the low-level features to reduce the redundancy of repetitive computation. Hence, it is able to use the memory more efficiently than conventional networks.

Moreover, it could mimic the ensemble learning with single network using the multiple probability vectors.

Since it is well known that the ensemble of multiple networks improves the performance, our proposed network has improved the performance of conventional residual network with 200 layers. However, since our network uses more parameters and the performance from ensemble of two independent residual networks is not reported, it is still open question that is it more efficient in number of parameters. Also, analyzing and resolving the correlation between the branches would be the future work.

This work has the benefit of reducing the memory requirement and convergence speed in deep networks. In addition, branched networks have advantages in terms of learning speed compared to conventional ensemble, which should learn several networks.

Other than the classification task, our proposed method is applicable to various vision tasks, which utilize ensemble for better performance. We expect our work to be useful in a variety of tasks.

## Acknowledgement

This work is supported by Hanwha Techwin CO., LTD.

## Reference

[1] A. Krizhevsky, I. Sutskever, G. E. Hinton, "ImageNet Classification with Deep Convolutional Neural Networks," Adv Neural Inf Processing Syst. 2012: 1097-1105.

[2] J. Deng, W. Dong, R. Socher, L.-J. Li, K. Li and L. Fei-Fei, "ImageNet: A Large-Scale Hierarchical Image Database," IEEE Computer Vision and Pattern Recognition (CVPR), 2009.

[3] C. Szegedy, V. Vanhoucke, S. Ioffe, J. Shlens, and Z. Wonjna, "Rethinking the Inception Architecture for Computer Vision," arXiv preprint arXiv:1512/00567 (2015).

[4] K. He, X. Zhang, S. Ren, and J. Sun, "Deep Residual Learning for Image Recognition, " IEEE Computer Vision and Pattern Recognition (CVPR), 2016.

[5] Jia, Yangqing and Shelhamer, Evan and Donahue, Jeff and Karayev, Sergey and Long, Jonathan and Girshick, Ross and Guadarrama, Sergio and Darrell, Trevor, "Caffe: Convolutional Architecture for Fast Feature Embedding", arXiv preprint arXiv:408.5093, (2014).

[6] G. Hinton, O. Vinyals, and J. Dean, "Distilling the Knowledge in a Neural network," Deep Learning and Representation Learning Workshop, NIPS, (2014).